\documentclass[11pt,a4paper]{article}
\usepackage[hyperref]{naaclhlt2019}
\usepackage{times}
\usepackage{latexsym}
\usepackage{graphicx}
\usepackage{subcaption}

\usepackage{url}

\aclfinalcopy 
\def\aclpaperid{52} 

\title{Plan, Write, and Revise: an Interactive System \\for Open-Domain Story Generation}

\author{Seraphina Goldfarb-Tarrant$^1$$^2$,  Haining Feng$^1$,  Nanyun Peng$^1$ \\
  $^1$ Information Sciences Institute, University of Southern California \\
  $^2$ University of Washington \\
  {\tt serif@uw.edu},
  {\tt haining@usc.edu}, 
  {\tt npeng@isi.edu} \\}

\begin{document}
\maketitle
\begin{abstract}
  Story composition is a challenging problem for machines and even for humans. We present a neural narrative generation system that interacts with humans to generate stories. Our system has different levels of human interaction, which enables us to understand at what stage of story-writing human collaboration is most productive, both to improving story quality and human engagement in the writing process. We compare different varieties of interaction in \textit{story-writing}, \textit{story-planning}, and \textit{diversity controls} under time constraints, and show that increased types of human collaboration at both planning and writing stages results in a 10-50\% improvement in story quality as compared to less interactive baselines. We also show an accompanying increase in user engagement and satisfaction with stories as compared to our own less interactive systems and to previous turn-taking approaches to interaction. Finally, we find that humans tasked with collaboratively improving a particular characteristic of a story are in fact able to do so, which has implications for future uses of human-in-the-loop systems.
\end{abstract}

\section{Introduction}
Collaborative human-machine story-writing has had a recent resurgence of attention from the research community \cite{roemmele2017eval,clark2018mil}. 
It represents a frontier for AI research; as a research community we have developed convincing NLP systems for some generative tasks like machine translation, but lag behind in creative areas like open-domain storytelling. Collaborative open-domain storytelling incorporates human interactivity for one of two aims: to improve human creativity via the aid of a machine, or to improve machine quality via the aid of a human. Previously existing approaches treat the former aim, and have shown that storytelling systems are not yet developed enough to help human writers. We attempt the latter, with the goal of investigating at what stage human collaboration is most helpful.

\newcite{gordon2009sayanything} use an information retrieval based system to write by alternating turns between a human and their system. \newcite{clark2018mil} use a similar turn-taking approach to interactivity, but employ a neural model for generation and allow the user to edit the generated sentence before accepting it. They find that users prefer a full-sentence collaborative setup (vs. shorter fragments) but are mixed with regard to the system-driven approach to interaction. \newcite{roemmele2017eval} experiment with a user-driven setup, where the machine doesn't generate until the user requests it to, and then the user can edit or delete at will. They leverage user-acceptance or rejection of suggestions as a tool for understanding the characteristics of a helpful generation. All of these systems involve the user in the \textit{story-writing} process, but lack user involvement in the \textit{story-planning} process, and so they lean on the user's ability to knit a coherent overall story together out of locally related sentences. They also do not allow a user to control the novelty or ``unexpectedness'' of the generations, which \newcite{clark2018mil} find to be a weakness. Nor do they enable iteration; a user cannot revise earlier sentences and have the system update later generations. We develop a system\footnote{The live demo is at \url{http://cwc-story.isi.edu}, with a video at \url{https://youtu.be/-hGd2399dnA}. Code and models are available at \url{https://github.com/seraphinatarrant/plan-write-revise}.} that allows a user to interact in all of these ways that were limitations in previous systems; it enables involvement in planning, editing, iterative revising, and control of novelty. We conduct experiments to understand which types of interaction are most effective for improving stories and for making users satisfied and engaged.  

We have two main interfaces that enable human interaction with the computer. There is \textit{cross-model} interaction, where the machine does all the composition work, and displays three different versions of a story written by three distinct models for a human to compare. The user guides generation by providing a topic for story-writing and by tweaking decoding parameters to control novelty, or \textit{diversity}. The second interface is \textit{intra-model} interaction, where a human can select the model to interact with (potentially after having chosen it via \textit{cross-model}), and can collaborate at all stages to jointly create better stories. The full range of interactions available to a user is: select a model, provide a topic, change diversity of content, collaborate on the planning for the story, and collaborate on the story sentences. It is entirely user-driven, as the users control how much is their own work and how much is the machine's at every stage. It supports revision; a user can modify an earlier part of a written story or of the story plan at any point, and observe how this affects later generations.
\section{System Description}
\label{sec:system}

\subsection{System Overview}
\begin{figure}
 \includegraphics[width=0.48\textwidth]{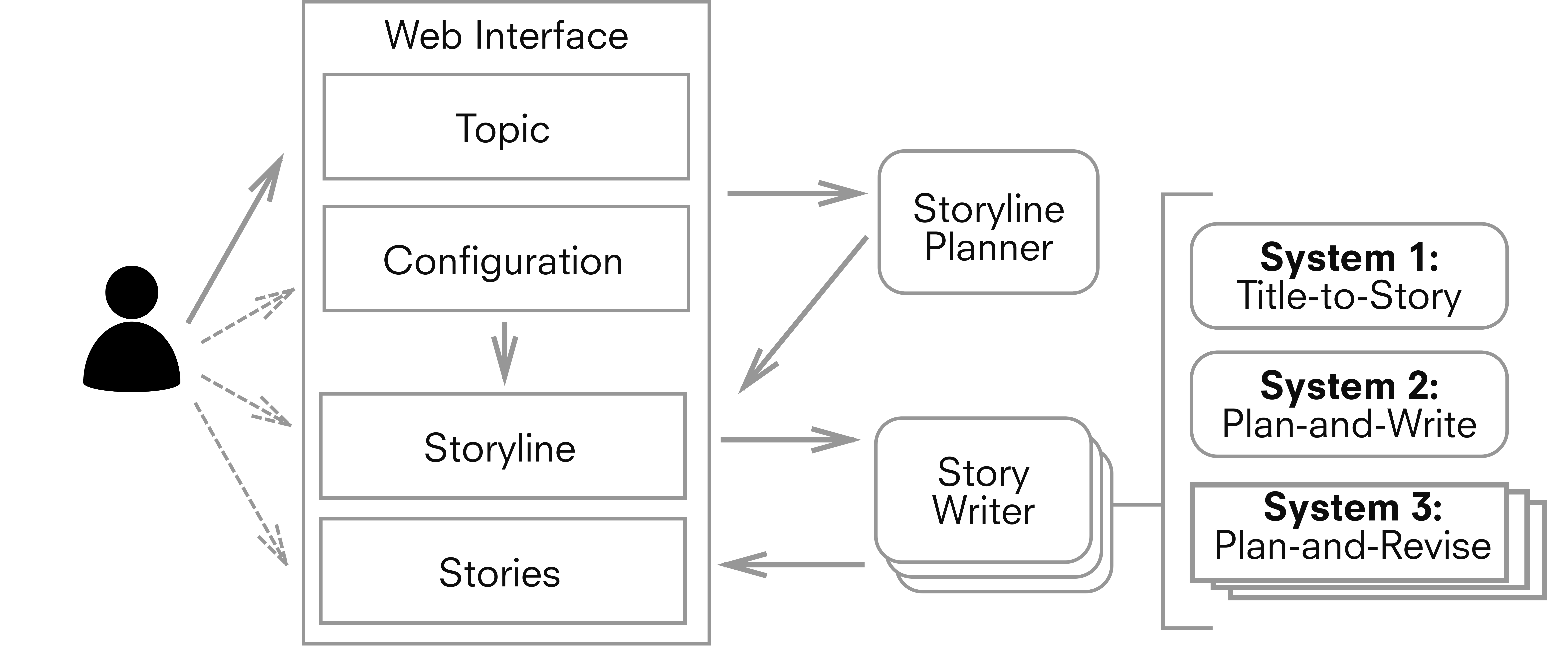} 
  \caption{Diagram of human-computer interaction mediated by the the demo system. The  dotted  arrows  represent optional interactions that the user can take. Depending on the set-up, the user may choose to interact with one or all story models.}
  \label{fig:system_diagram}
  \vspace{-1.5em}
\end{figure}
Figure \ref{fig:system_diagram} shows a diagram of the interaction system. The dotted arrows represent optional user interactions. 
\paragraph{Cross-model mode} requires the user to enter a \textit{topic}, such as ``the not so haunted house'', and can optionally vary the \textit{diversity} used in the \textsc{Storyline Planner} or the \textsc{Story Writer}. \textit{Diversity} numbers correspond directly to softmax temperatures, which we restrict to a reasonable range, determined empirically. The settings are sent to the \textsc{Storyline Planner} module, which generates a storyline for the story in the form of a sequence of phrases as per the method of \newcite{yao2018plan}. Everything is then sent to the \textsc{Story Writer}, which will return three stories. 

\paragraph{Intra-model mode} enables advanced interactions with one story system of the user's choice. The \textsc{Storyline Planner} returns either one storyline phrase or many, and composes the final storyline out of the combination of phrases the system generated, the user has written, and edits the user has made. 
These are sent to the \textsc{Story Writer}, which returns either a single sentence or a full story as per user's request. The process is flexible and iterative. The user can choose how much or little content they want to provide, edit, or re-generate, and they can return to any step at any time until they decide they are done.

\paragraph{Pre-/Post-processing and OOV handling} To enable interactive flexibility, the system must handle open-domain user input. User input is lower-cased and tokenized to match the model training data via spaCy\footnote{\url{spacy.io}}. Model output is naively detokenized via Moses \cite{koehn2007moses} based on feedback from users that this was more natural. 
User input OOV handling is done via WordNet \cite{wordnet} by recursively searching for hypernyms and hyponyms (in that order) until either an in-vocabulary word is found or until a maximum distance from the initial word is reached.\footnote{\textit{distance} is difference og levels in the WordNet hierarchy, and was set empirically to 10.} We additionally experimented with using cosine similarity to GloVe vectors \cite{pennington2014glove}, but found that to be slower and not qualitatively better for this domain. 

\subsection{Web Interface}
\begin{figure*}[t!]
    \centering
    \begin{subfigure}[t]{0.44\textwidth}
        \includegraphics[width=\linewidth,height=11.5em]{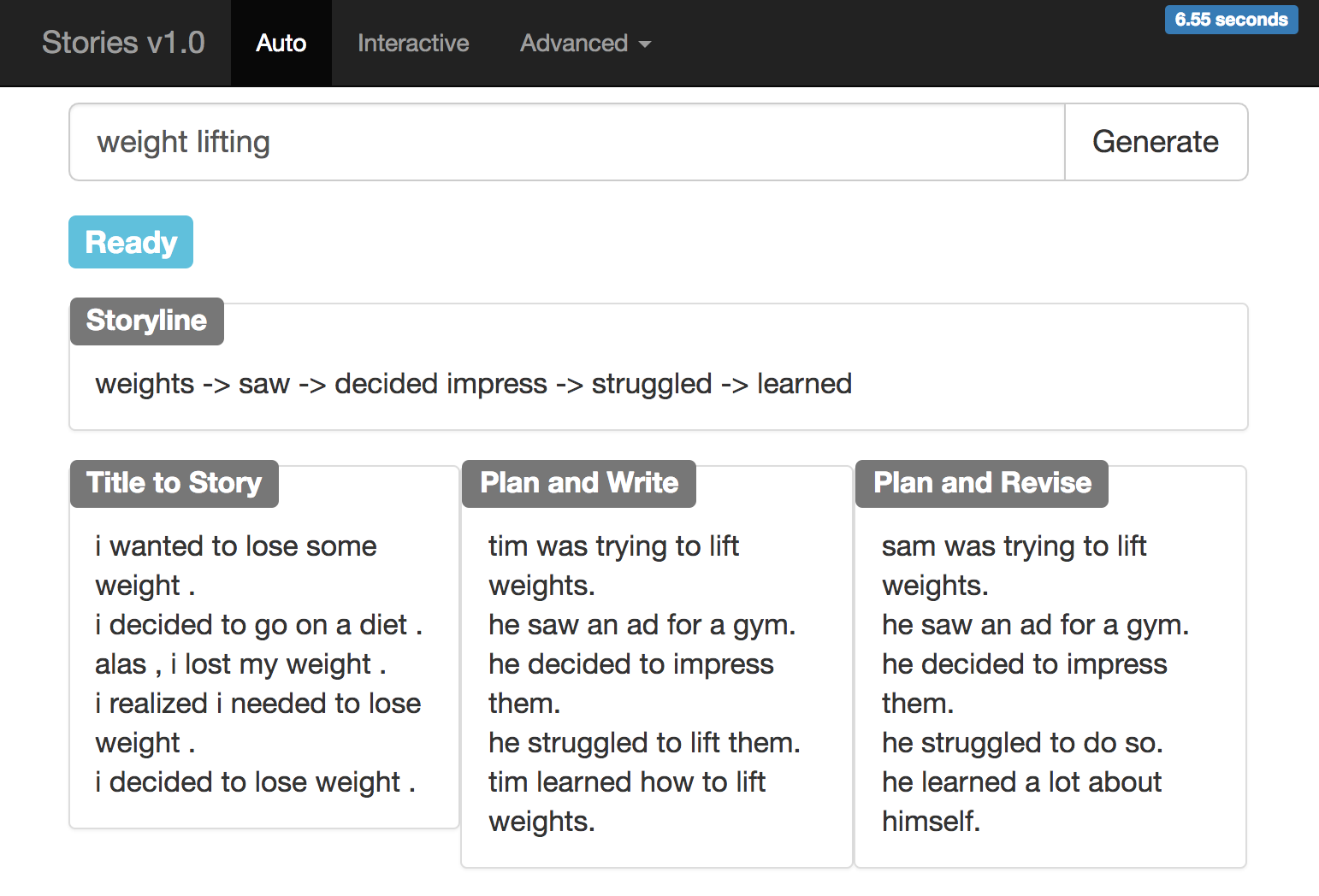}
        \caption{cross-model interaction, comparing three models with advanced options to alter the storyline and story diversities.}
        \label{fig:auto_screenshot}
    \end{subfigure} \quad
    \begin{subfigure}[t]{0.49\textwidth}
        \includegraphics[width=\linewidth,height=11.5em]{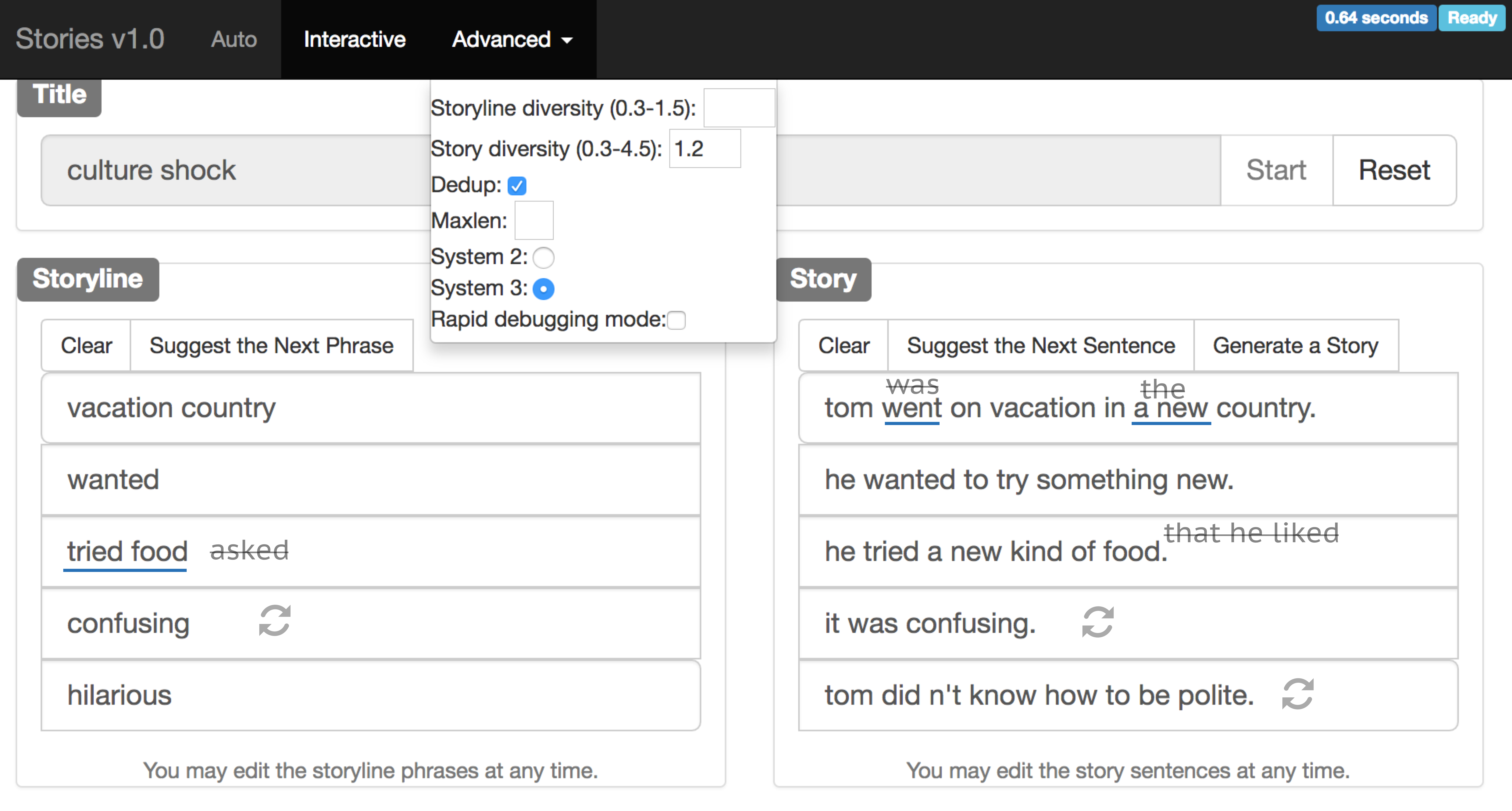}
        \caption{intra-model interaction, showing advanced options and annotated with user interactions from an example study.}
        \label{fig:interactive_screenshot}
    \end{subfigure}
    \vspace{-1em}
    \caption{Screenshots of the demo user interface}\label{fig:ui_screenshots}
    \vspace{-1em}
\end{figure*}
Figure \ref{fig:ui_screenshots} shows screenshots for both the \textit{cross-model} and \textit{intra-model} modes of interaction. Figure~\ref{fig:auto_screenshot} shows that the \textit{cross-model} mode makes clear the differences between different model generations for the same topic. Figure~\ref{fig:interactive_screenshot} shows the variety of interactions a user can take in \textit{intra-model} interaction, and is annotated with an example-in-action. User inserted text is underlined in blue, generated text that has been removed by the user is in grey strike-through. The \textit{refresh} symbol marks areas that the user re-generated to get a different sentence (presumably after being unhappy with the first result). As can be seen in this example, minor user involvement can result in a significantly better story. 

\subsection{Model Design}
All models for both the \textsc{Storyline Planner} and \textsc{Story Writer} modules are conditional language models implemented with LSTMs based on \newcite{merity2018regularizing}. These are 3-stacked LSTMs that include weight-dropping, weight-tying, variable length back propagation with learning rate adjustment, and Averaged Stochastic Gradient Descent (ASGD). They are trained on the ROC dataset \cite{mostafazadeh2016corpus}, which after lowercasing and tokenization has a vocabulary of 38k. Storyline Phrases are extracted as in \newcite{yao2018plan} via the RAKE algorithm \cite{rose2010automatic} which results in a slightly smaller Storyline vocabulary of 31k. The \textsc{Storyline Planner} does decoding via sampling to encourage creative exploration. The \textsc{Story Writer} has an option to use one or all three systems, all of which decode via beamsearch and are detailed below.

The \textit{Title-to-Story} system is a baseline, which generates directly from topic. 

The \textit{Plan-and-Write} system adopts the static model in \newcite{yao2018plan} to use the storyline to supervise story-writing.

\textit{Plan-and-Revise} is a new system that combines the strengths of \newcite{yao2018plan} and \newcite{holtzman2018learning}. It supplements the Plan-and-Write model by training two discriminators on the ROC data and using them to re-rank the LSTM generations to prefer increased {\em creativity} and {\em relevance}.\footnote{\newcite{holtzman2018learning} use four discriminators, but based on ablation testing we determined these two to perform best on our dataset and for our task.} Thus the decoding objective of this system becomes 
$ f_{\lambda}(x,y) = log(P_{lm}(y|x)) + \sum_k \lambda_{k}s_{k}(x,y) $ where $P_{lm}$ is the conditional language model probability of the LSTM, $s_k$ is the discriminator scoring function, and $\lambda_k$ is the learned weight of that discriminator. At each timestep all live beam hypotheses are scored and re-ranked. Discriminator weights are learnt by minimizing Mean Squared Error on the difference between the scores of gold standard and generated story sentences.

\section{Experiments}
\label{sec:experiments}
We experiment with six types of interaction: five variations created by restricting different capabilities of our system, and a sixth turn-taking baseline that mimics the interaction of the previous work~\cite{clark2018mil,gordon2009sayanything}. We choose our experiments to address the research questions: What type of interaction is most engaging? Which type results in the best stories? Can a human tasked with correcting for certain weaknesses of a model successfully do so? The variations on interactions that we tested are:
\begin{enumerate}
\itemsep-.3em 
    \item Machine only: no human-in-loop.
    \item Diversity only: user can compare and select models but only diversity is modifiable.
    \item Storyline only: user collaborates on storyline but not story.
    \item Story only: user collaborates on story but not storyline.
    \item All: user can modify everything.
    \item Turn-taking: user and machine take turns writing a sentence each (user starts). user can edit the machine-generations, but once they move on to later sentences, previous sentences are read-only.\footnote{This as closely matches the previous work as possible with our user interface. This model does not use a storyline.}
\end{enumerate}
We expand experiment 5 to answer the question of whether a human-in-the-loop interactive system can address specific shortcomings of generated stories. We identify three types of weaknesses common to generation systems -- \textit{Creativity}, \textit{Relevance}, and \textit{Causal \& Temporal Coherence}, and conduct experiments where the human is instructed to focus on improving specifically one of them. The targeted human improvement areas intentionally match the \textit{Plan-and-Revise} discriminators, so that, if successful, the "human discriminator" data can assist in training the machine discriminators. All experiments (save experiment 2, which lets the user pick between models) use the \textit{Plan-and-Revise} system.
\subsection{Details}
We recruit 30 Mechanical Turk workers per experiment (270 unique workers total) to complete story writing tasks with the system.\footnote{We enforce uniqueness to prevent confounding effects from varying levels of familiarity with the demo UI} We constrain them to ten minutes of work (five for writing and five for a survey) and provide them with a fixed \textit{topic} to control this factor across experiments. They co-create a story and complete a questionnaire which asks them to self-report on their engagement, satisfaction, and perception of story quality.\footnote{Text of questionnaire and other Mechanical Turk materials are included in Appendix \ref{sec:user_study}} For the additional focused error-correction experiments, we instruct Turkers to try to improve the machine-generated stories with regard to the given aspect, under the same time constraints. As an incentive, they are given a small bonus if they are later judged to have succeeded. 

We then ask a separate set of Turkers to rate the stories for overall quality and the three improvement areas. All ratings are on a five-point scale. We collect two ratings per story, and throw out ratings that disagree by more than 2 points. 
A total of 11\% of ratings were thrown out, leaving four metrics across 241 stories for analysis. 

\begin{table}
\small
  \centering
    \begin{tabular}{l|c|c|c|c}
    \hline \bf Experiment & \bf E & \bf Q & \bf S & \bf Use Again \\
    \hline
    Diversity only & 3.77 & 2.90 & 3.27 & 1.40 \\
    Storyline only & 4.04 & 3.36 & 3.72 & 1.27 \\
    Story only & \bf 4.50 & 3.17 & 3.60 & 1.60 \\
    All & 4.41 & 3.55 & 3.76 & 1.55 \\
    All + Creative & 4.00 & 3.27 & 3.70 & \bf 1.70 \\
    All + Relevant & 4.20 & 3.47 & 3.83 & 1.57 \\
    All + C-T & 4.30 & \bf 3.77 & \bf 4.30 & 1.53 \\
    Turn-taking & 4.31 & 3.38 & 3.66 & 1.52 \\
    \hline
    \end{tabular}
  \caption{\label{tab:self-report} User self-reported scores, from 1-5. E: Entertainment value, Q: Quality of Story, S: Satisfaction with Story.  Note that the final column \textit{Use Again } is based on converting ``no'' to 0, ``conditional'' to 1, and ``yes'' to 2.}
\end{table}

\begin{table}
\small
\begin{tabular}{l@{ }|c@{ }|c@{ }|c@{ }|c@{ }}
\hline \bf Experiment & \bf Overall & \bf Creative & \bf Relevant & \bf C-T \\ 
\hline
Machine & 2.34\phantom{*} & 2.68\phantom{*} & 2.46\phantom{*} & 2.54\phantom{*} \\
Diversity only & 2.50\phantom{*} & 2.96\phantom{*} & 2.75\phantom{*} & 2.81\phantom{*} \\
Storyline only & 3.21\phantom{*} & 3.27\phantom{*} & 3.88\phantom{*} & 3.65\phantom{*} \\
Story only & 3.70$^{*}$ & \textbf{4.04}$^{*}$ & 3.96$^{*}$ & \textbf{4.24}$^{*}$ \\
All & 3.54\phantom{*} & 3.62\phantom{*} & 3.93$^{*}$ & 3.83\phantom{*} \\
All + Creative & \textbf{3.73}$^{*}$ & 3.96$^{*}$ & 3.98$^{*}$ & 3.93$^{*}$ \\
All + Relevant & 3.53$^{*}$  & 3.52\phantom{*} & 4.05$^{*}$ & 3.91$^{*}$ \\
All + C-T & 3.62$^{*}$ & 3.88$^{*}$ & 4.00$^{*}$ & 3.98$^{*}$ \\
Turn-taking & 3.55$^{*}$ & 3.68\phantom{*} & \textbf{4.27}$^{*}$ & 3.81\phantom{*} \\
\hline
\end{tabular}
\caption{\label{tab:results} Results for all experiments, from 1-5. Best scores per metric are bolded, scores not significantly different ($\alpha=0.1$, per Wilcoxon Signed-Rank Test) are starred. C-T stands for Causal-Temporal Coherence, the \textbf{+} experiments are the extensions where the user focuses on improving a particular quality.}
\vspace{-1em}
\end{table}

\section{Results} 
\label{sec:results}

\paragraph{User Engagement}
Self-reported scores are relatively high across the board, as can be seen in Table \ref{tab:self-report}, with the majority of users in all experiments saying they would like to use the system again. The lower scores in the {\em Diversity only} and {\em Storyline only} experiments are elucidated by qualitative comments from users of frustration at the inability to sufficiently control the generations with influence over only those tools. \textit{Storyline only} is lowest for \textit{Use Again}, which can be explained by the model behavior when dealing with unlikely storyline phrases. Usually, the most probable generated story will contain all storyline phrases (exact or similar embeddings) in order, but there is no mechanism that strictly enforces this. When a storyline phrase is uncommon, the story model will often ignore it. Many users expressed frustration at the irregularity of their ability to guide the model when collaborating on the storyline, for this reason. 

Users were engaged by collaboration; all experiments received high scores on being entertaining, with the collaborative experiments rated more highly than {\em Diversity only}. The pattern is repeated for the other scores, with users being more satisfied and feeling their stories to be higher quality for all the more interactive experiments. The \textit{Turn-taking} baseline fits into this pattern; users prefer it more than the less interactive \textit{Diversity only} and \textit{Storyline only}, but often (though not always) less than the more interactive \textit{Story only, All, All+} experiments. Interestingly, user perception of the quality of their stories does not align well with independent rankings. Self-reported quality is low in the {\em Story only} experiment, which contrasts with it being highest rated independently (as discussed below). Self-reported scores also suggest that users judge their stories to be much better when they have been focusing on causal-temporal coherence, though this focus carries over to a smaller improvement in independent rankings. While it is clear that additional interactivity is a good idea, the disjunct between user perception of their writing and reader perception under different experiment conditions is worthwhile to consider for future interactive systems. 

\paragraph{Story Quality}
As shown in Table \ref{tab:results}, human involvement of \textit{any kind} under tight constraints helps story quality across all metrics, with mostly better results the more collaboration is allowed. The exception to this trend is \textit{Story only} collaboration, which performs best or close to best across the board. This was unexpected; it is possible that these users benefited from having to learn to control only \textit{one} model, instead of both, given the limited time. It is also possible that being forced to be reliant on system storylines made these users more creative. 

\paragraph{Turn-taking Baseline}
The turn-taking baseline performs comparably in overall quality and relevance to other equally interactive experiments (\textit{Story only}, \textit{All, All+}). It achieves highest scores in relevance, though the top five systems for relevance are not statistically significantly different. It is outperformed on creativity and causal-temporal coherence by the strong \textit{Story only} variation, as well as the \textit{All, All+} systems. This suggests that local sentence-level editing is sufficient to keep a story on topic and to write well, but that creativity and causal-temporal coherence require some degree of global cohesion that is assisted by iterative editing. The same observation as to the strength of \textit{Story only} over \textit{All} applies here as well; turn-taking is the least complex of the interactive systems, and may have boosted performance from being simpler since time was constrained and users used the system only once. Thus a turn-based system is a good choice for a scenario where users use a system infrequently or only once, but the comparative performance may decrease in future experiments with more relaxed time constraints or where users use the system repeatedly.

\paragraph{Targeted Improvements}
The results within the \textit{All} and \textit{All +} setups confirm that stories can be improved with respect to a particular metric. The diagonal of strong scores displays this trend, where the creativity-focused experiment has high creativity, etc. An interesting side effect to note is that focusing on \textit{anything} tends to produce better stories, reflected by higher overall ratings. \textit{All + Relevance} is an exception which does not help creativity or overall (perhaps because relevance instantly becomes very high as soon a human is involved), but apart from that \textit{All +} experiments are better across all metrics than \textit{All}. This could mean a few things: that when a user improves a story in one aspect, they improve it along the other axes, or that users reading stories have trouble rating aspects entirely independently.

\section{Conclusions and Future Work}
\label{sec:conclusion}
We have shown that all levels of human-computer collaboration improve story quality across all metrics, compared to a baseline computer-only story generation system. We have also shown that flexible interaction, which allows the user to return to edit earlier text, improves the specific metrics of creativity and causal-temporal coherence above previous rigid turn-taking approaches. We find that, as well as improving story quality, more interaction makes users more engaged and likely to use the system again. Users tasked with collaborating to improve a specific story quality were able to do so, as judged by independent readers. 

As the demo system has successfully used an ensemble of collaborative discriminators to improve the same qualities that untrained human users were able to improve even further, this suggests promising future research into human-collaborative stories as training data for new discriminators. It could be used both to strengthen existing discriminators and to develop novel ones, since discriminators are extensible to arbitrarily many story aspects.
\section*{Acknowledgments}
We thank the anonymous reviewers for their feedback, as well as the members of the PLUS lab for their thoughts and iterative testing.
This work is supported by Contract W911NF-15-1-0543 with the US Defense Advanced Research Projects
Agency (DARPA). \\
\clearpage

\bibliography{naaclhlt2019}

\begin{thebibliography}{11}
\expandafter\ifx\csname natexlab\endcsname\relax\def\natexlab#1{#1}\fi

\bibitem[{Clark and Smith(2018)}]{clark2018mil}
Anne Spencer Ross Chenhao Tan Yangfeng~Ji Clark, Elizabeth and Noah~A. Smith.
  2018.
\newblock Creative writing with a machine in the loop: Case studies on slogans
  and stories.
\newblock In \emph{23rd International Conference on Intelligent User Interfaces
  (IUI)}.

\bibitem[{Holtzman et~al.(2018)Holtzman, Buys, Forbes, Bosselut, Golub, and
  Choi}]{holtzman2018learning}
Ari Holtzman, Jan Buys, Maxwell Forbes, Antoine Bosselut, David Golub, and
  Yejin Choi. 2018.
\newblock Learning to write with cooperative discriminators.
\newblock In \emph{Proceedings of ACL}.

\bibitem[{Koehn et~al.(2007)Koehn, Hoang, Birch, Callison-Burch, Federico,
  Bertoldi, Cowan, Shen, Moran, Zens et~al.}]{koehn2007moses}
Philipp Koehn, Hieu Hoang, Alexandra Birch, Chris Callison-Burch, Marcello
  Federico, Nicola Bertoldi, Brooke Cowan, Wade Shen, Christine Moran, Richard
  Zens, et~al. 2007.
\newblock Moses: Open source toolkit for statistical machine translation.
\newblock In \emph{Proceedings of the 45th annual meeting of the ACL on
  interactive poster and demonstration sessions}, pages 177--180. Association
  for Computational Linguistics.

\bibitem[{Merity et~al.(2018)Merity, Keskar, and
  Socher}]{merity2018regularizing}
Stephen Merity, Nitish~Shirish Keskar, and Richard Socher. 2018.
\newblock Regularizing and optimizing lstm language models.
\newblock In \emph{Proceedings of the Sixth International Conference on
  Learning Representations (ICLR)}.

\bibitem[{Miller(1995)}]{wordnet}
George~A. Miller. 1995.
\newblock Wordnet: A lexical database for english.
\newblock In \emph{Communications of the ACM Vol. 38}.

\bibitem[{Mostafazadeh et~al.(2016)Mostafazadeh, Chambers, He, Parikh, Batra,
  Vanderwende, Kohli, and Allen}]{mostafazadeh2016corpus}
Nasrin Mostafazadeh, Nathanael Chambers, Xiaodong He, Devi Parikh, Dhruv Batra,
  Lucy Vanderwende, Pushmeet Kohli, and James Allen. 2016.
\newblock A corpus and cloze evaluation for deeper understanding of commonsense
  stories.
\newblock In \emph{Proceedings of NAACL-HLT}, pages 839--849.

\bibitem[{Pennington et~al.(2014)Pennington, Socher, and
  Manning}]{pennington2014glove}
Jeffrey Pennington, Richard Socher, and Christopher~D. Manning. 2014.
\newblock \href {http://www.aclweb.org/anthology/D14-1162} {Glove: Global
  vectors for word representation}.
\newblock In \emph{Empirical Methods in Natural Language Processing (EMNLP)},
  pages 1532--1543.

\bibitem[{Roemmele and Swanson.(2017)}]{roemmele2017eval}
Andrew S.~Gordon Roemmele, Melissa and Reid Swanson. 2017.
\newblock Evaluating story generation systems using automated linguistic
  analyses.
\newblock In \emph{SIGKDD 2017 Workshop on Machine Learning for Creativity}.

\bibitem[{Rose et~al.(2010)Rose, Engel, Cramer, and Cowley}]{rose2010automatic}
Stuart Rose, Dave Engel, Nick Cramer, and Wendy Cowley. 2010.
\newblock Automatic keyword extraction from individual documents.
\newblock In \emph{Text Mining: Applications and Theory}.

\bibitem[{Swanson and Gordon(2009)}]{gordon2009sayanything}
Reid Swanson and Andrew~S. Gordon. 2009.
\newblock Say anything: A demonstration of open domain interactive digital
  storytelling.
\newblock In \emph{Joint International Conference on Interactive Digital
  Storytelling}.

\bibitem[{Yao et~al.(2019)Yao, Peng, Ralph, Knight, Zhao, and
  Yan}]{yao2018plan}
Lili Yao, Nanyun Peng, Weischedel Ralph, Kevin Knight, Dongyan Zhao, and Rui
  Yan. 2019.
\newblock Plan-and-write: Towards better automatic storytelling.
\newblock In \emph{Proceedings of the thirty-third AAAI Conference on
  Artificial Intelligence (AAAI)}.

\end{thebibliography}
\bibliographystyle{acl_natbib}

\clearpage

\appendix
\section{Demo Video}
The three-minute video demonstrating the interaction capabilities of the system can be viewed at \url{https://youtu.be/-hGd2399dnA}. (Same video as linked in the paper footnote).

\section{Training and Decoding Parameters}
\label{sec:hyper_params}
\subsection{Decoding}
\label{sec:decoding}
Default diversity (Softmax Temperature) for Storyline Planner is \textit{0.5}, for Story Writer it is \textit{None} (as beamsearch is used an thus can have but does not require a temperature). Beam size for all Story Writer models is \textit{5}. Additionally, Storyline Phrases are constrained to be unique (unless a user duplicates them), and Beamsearch is not normalized by length (both choices determined empirically).
\subsection{Training}
\label{sec:training}
We follow the parameters used in \newcite{yao2018plan} and \newcite{merity2018regularizing}.\\ 
\begin{table}[h]
\small
\begin{tabular}{l@{}|c@{}|c}
\hline
\bf Parameter & Storyline Model  & Story Models  \\ 
\hline
Embedding Dim & 500 & 1000  \\
\hline
Hidden Layer Dim & 1000 & 1500 \\
\hline
Input Embedding Dropout & 0.4 & 0.2 \\
\hline
Hidden Layer Dropout & 0.1 & 01 \\
\hline
Batch Size & 20 & 20 \\
\hline
BPTT & 20 & 75 \\
\hline
Learning Rate & 10 & 10 \\
\hline
Vocabulary size & 31,382 & 37,857 \\
\hline
Total Model Parameters & 32,489,878 & 80,927,858 \\
\hline
Epochs & 50 & 120 \\
\hline
\end{tabular} 
\caption{Training parameters for models used in demo.}\label{tab:train}
\end{table}

\section{User Study} 
\label{sec:user_study}
\subsection{Questionnaire}
\label{sec:questionnaire}

\begin{table}[h]
\small
\begin{tabular}{l}
\hline
{\bf Post Story Generation Questionnaire}\\ \hline
\textit{How satisfied are you with the final story?}  \\
\hline
\textit{What do you think is the overall quality of the final story?}  \\
\hline
\textit{Was the process entertaining?}   \\
\hline
\textit{Would you use the system again?}   \\
\hline
\end{tabular} 
\caption{Questionnaire for user self-reporting, range 1 to 5 (1 low).}\label{tab:survey}
\end{table}

\subsection{Mechanical Turk Materials}
\label{sec:mturk}
\begin{figure*}
  \caption{Template \& Instructions for Writing Stories in the \textit{All + Creative} experiment.}
  \centering
    \includegraphics[width=\textwidth]{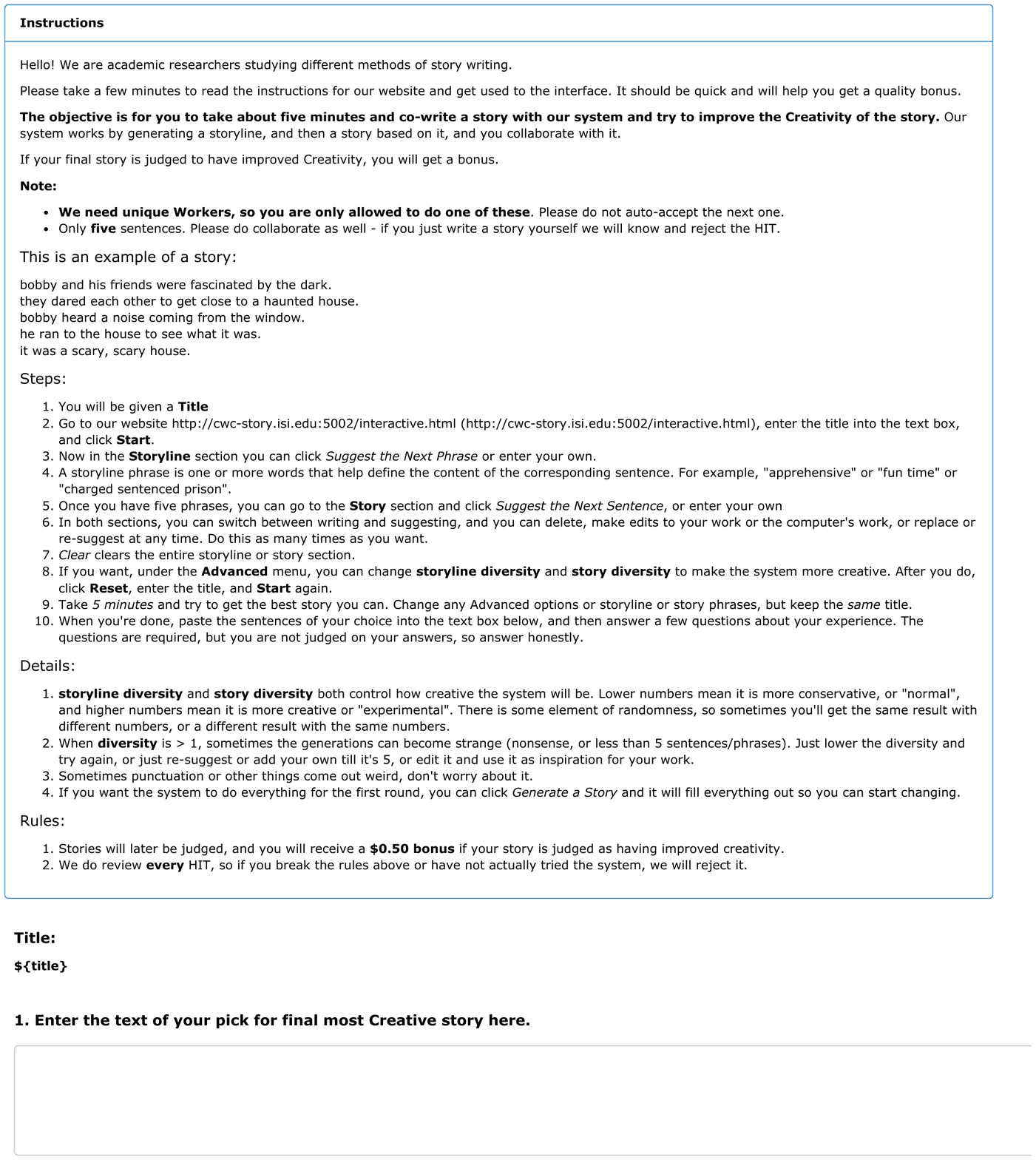}
    \label{fig:mturk_writing}
\end{figure*}

\begin{figure*}
  \caption{Template \& Instructions for Ranking Stories}
  \centering
    \includegraphics[width=\textwidth]{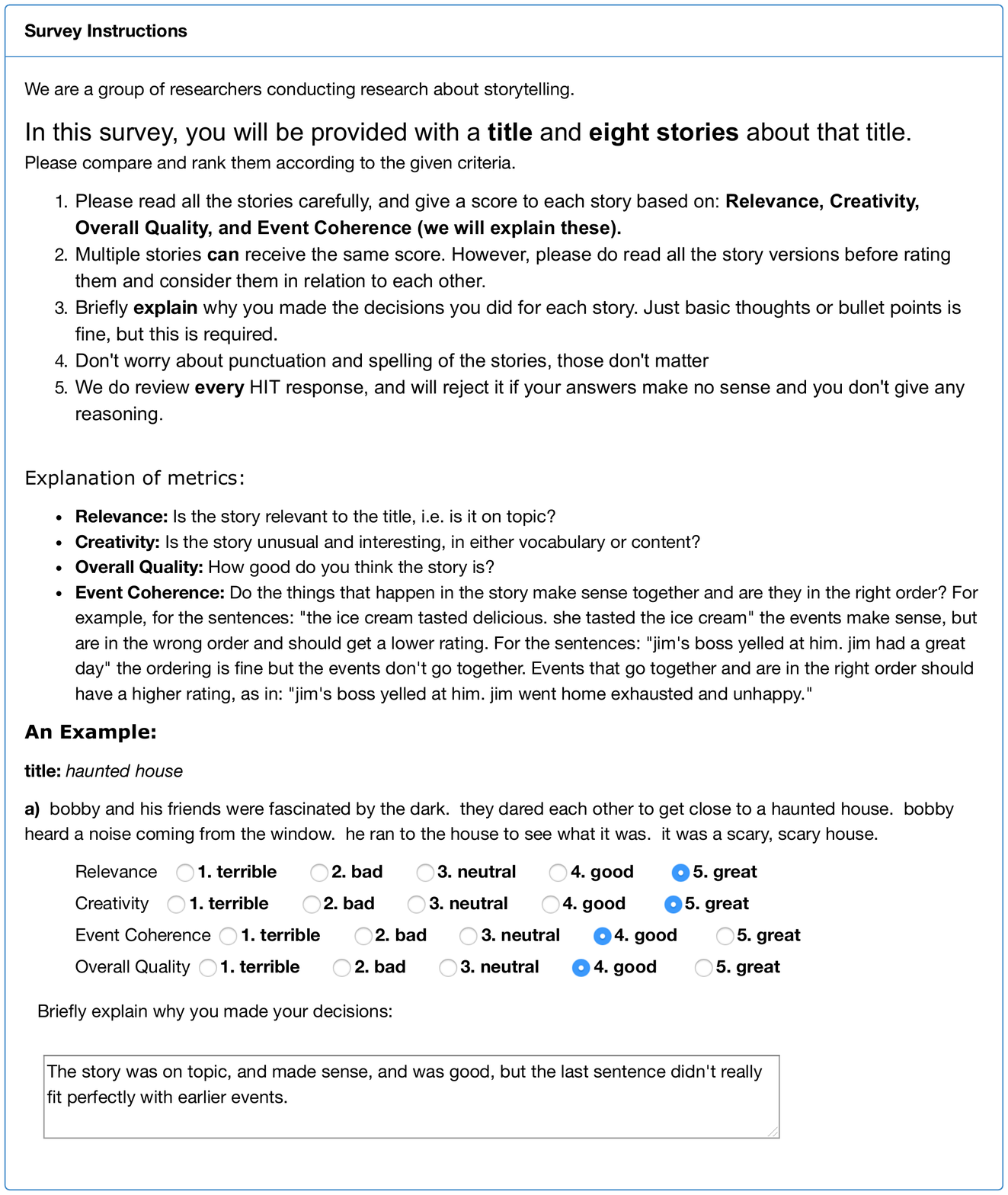}
    \label{fig:mturk_ranking}
\end{figure*}

Following are examples of the materials used in doing Mechanical Turk User Studies. Figure \ref{fig:mturk_writing} is an example of the \textit{All + Creative} focused experiment for \textit{story-writing}. The instructions per experiment differ across all, but the template is the same. Figure \ref{fig:mturk_ranking} is the survey for ranking stories across various metrics. This remains constant save that story order was shuffled every time to control for any effects of the order a story was read in.

\pagebreak

\end{document}